%% file: main.tex
\documentclass{preprint}
\usepackage{booktabs}
\usepackage{graphicx}
\usepackage[colorlinks=true]{hyperref}
\usepackage[numbers,comma,sort&compress,super]{natbib} 
\usepackage{xcolor}
\usepackage{makecell}
\usepackage{tabularx} 
\usepackage{enumitem}
\usepackage{subcaption}
\usepackage[LY1]{fontenc}
\DeclareTextCompositeCommand{\k}{LY1}{a}
  {\oalign{a\crcr\noalign{\kern-.27ex}\hidewidth\char7}}

\title{Extracting Post-Acute Sequelae of SARS-CoV-2 Infection Symptoms from Clinical Notes via Hybrid Natural Language Processing}

\author[1]{Zilong Bai}
\author[1]{Zihan Xu}
\author[1]{Cong Sun}
\author[1]{Chengxi Zang}
\author[2]{H. Timothy Bunnell}
\author[1]{Catherine Sinfield}
\author[3]{Jacqueline Rutter}
\author[3]{Aaron Thomas Martinez}
\author[4]{L. Charles Bailey}
\author[1]{Mark Weiner}
\author[1]{Thomas R. Campion}
\author[5]{Thomas Carton}
\author[4]{Christopher B. Forrest}
\author[1]{Rainu Kaushal}
\author[1]{Fei Wang}
\author[1,*]{Yifan Peng}
\affil[1]{Population Health Sciences, Weill Cornell Medicine, New York, USA.}
\affil[2]{Nemours Children's Health, Wilmington, USA.}
\affil[3]{RECOVER Patient, Caregiver, or Community Advocate Representative, New York, USA.}
\affil[4]{Applied Clinical Research Center, Children's Hospital of Philadelphia, Philadelphia, USA.}
\affil[5]{Louisiana Public Health Institute, New Orleans, USA.}
\affil[*]{Corresponding author(s). Email(s): \url{yip4002@med.cornell.edu}}

\begin{document}

\maketitle

\begin{abstract}
Accurately and efficiently diagnosing Post-Acute Sequelae of COVID-19 (PASC) remains challenging due to its myriad symptoms that evolve over long- and variable-time intervals. To address this issue, we developed a hybrid natural language processing pipeline that integrates rule-based named entity recognition with BERT-based assertion detection modules for PASC-symptom extraction and assertion detection from clinical notes. We developed a comprehensive PASC lexicon with clinical specialists. From 11 health systems of the RECOVER initiative network across the U.S., we curated 160 intake progress notes for model development and evaluation, and collected 47,654 progress notes for a population-level prevalence study. We achieved an average F1 score of 0.82 in one-site internal validation and 0.76 in 10-site external validation for assertion detection. Our pipeline processed each note at $2.448\pm 0.812$ seconds on average. Spearman correlation tests showed $\rho >0.83$ for positive mentions and $\rho >0.72$ for negative ones, both with $P <0.0001$. These demonstrate the effectiveness and efficiency of our models and their potential for improving PASC diagnosis.
\end{abstract}

% \begin{keywords}
% First keyword \and Second keyword \and More
% \end{keywords}

\section{INTRODUCTION}

Post-acute sequelae of coronavirus disease 2019 (PASC), or Long COVID, is an often debilitating and complex infection-associated chronic condition (IACC) that occurs after SARS-CoV-2 infection and is present for at least 3 months beyond the acute phase of the infection\cite{fineberg2024covid, thaweethai2023development}. PASC has affected tremendous individuals globally\cite{thaweethai2023development, Ford2023-mf, ballering2022persistence, Davis2023-vx}, with over 200 distinct symptoms reported across multiple organ systems, emphasizing its complexity and multifaceted nature\cite{Davis2023-vx, Michelen2021-bp, parums2023editorial}. Currently, diagnosing, treating, and caring for patients with PASC remains challenging due to its myriad symptoms that evolve over long- and variable-time intervals\cite{fineberg2024covid, Davis2023-vx, McCaddon2021-za, miller2023poll}. The accurate characterization and identification of PASC patients are critical for effectively managing this evolving public health issue, such as accurate diagnosis, stratification of risk among patients, evaluation of the impact of therapeutics and immunizations, and ensuring diverse recruitment for clinical research studies\cite{fineberg2024covid, Davis2023-vx}.

Electronic health records (EHRs) have been widely used in clinical practice and AI in healthcare research. Currently, EHR analyses regarding PASC predominantly rely on structured data, such as billing diagnoses commonly recorded as ICD-10 codes. Yet, existing diagnostic codes for PASC (i.e., U09.9) have been shown to lack the requisite sensitivity and specificity for an accurate PASC diagnosis\cite{oHare2022complexity, pfaff2023coding, duerlund2022positive, ioannou2022rates, zhang2023potential}. Additionally, findings reveal demographic biases among patients coded with U09.9, showing a higher prevalence in women, White, non-Hispanic individuals, and those from areas of lower poverty and higher education\cite{zhang2023potential, azhir2025precision}. Structured billing diagnoses have limitations in assessing the true frequency of conditions associated with PASC, such as Postural Orthostatic Tachycardia Syndrome (POTS), which is diagnosed in 2-14\% of individuals after a COVID-19 infection\cite{ormiston2022postural}. POTS did not have a specific ICD-10 code until October 1, 2022. Additionally, its primary symptoms--palpitations and dizziness--may be documented in clinical notes but often go unrecognized as part of the syndrome by clinicians, leading to underreporting in billing data. Comparable challenges are also found in diagnosing Myalgic Encephalomyelitis/Chronic Fatigue Syndrome (ME/CFS) when relying solely on structured data\cite{madhavan2021use, unger2024myalgic, dehlia2024persistence}. These biases and limitations complicate our understanding and management of PASC and highlight the need for improved diagnostic approaches.

Unstructured narratives, found in EHR notes, contain detailed accounts of a patient's clinical history, symptoms, and the effects of those symptoms on physical and cognitive functioning. Previous studies have applied NLP to detect acute COVID symptoms within free-form text notes\cite{Palmer2024-oo, McMurry2024-qn, Liu2023-pn, Hurley2023-ol}. Studies on acute COVID-19 symptom extraction have achieved promising results\cite{Canales2021-jt}. However, due to the inherent difference between PASC and COVID-19, such NLP approaches are not directly transferable\cite{McMurry2024-qn, Hurley2023-ol, Canales2021-jt}. Other research has concentrated on identifying PASC symptoms without assertion detection. The PASCLex study by Wang et al.\cite{Wang2022-ua} developed a lexicon of symptoms and synonyms, employing a rule-based approach to search for symptoms within clinical notes.

To address the challenges of identifying PASC and build on previous research to improve diagnostic methods, we developed a hybrid NLP pipeline that integrates rule-based and deep-learning approaches. This hybrid pipeline not only detects relevant symptoms in clinical notes but also accurately determines their assertion status--whether a symptom is truly present or non-present. Our approach was implemented within data from the RECOVER Initiative\cite{recover2024-tc}, which provides access to large, diverse COVID-19 and PASC patient populations with electronic health records (EHR) data from a network of large health systems across the United States. One of the key features of our pipeline is a comprehensive PASC lexicon, developed in collaboration with clinicians, consisting of 25 symptom categories and 798 Unified Medical Language System (UMLS) concepts for precise symptom identification. Additionally, we designed a BERT-based module specifically to assess symptom assertions, distinguishing between symptoms that are present or non-present (including absent, uncertain, or other possible statuses, as detailed in the Methods). To ensure our approach extracts PASC symptoms from unstructured clinical notes robustly and efficiently, we curated 60 progress notes from New York-Presbyterian/Weill Cornell Medicine (WCM) for model development and internal validation, and 100 progress notes from 10 additional sites for external validation. To strengthen our evaluation in comparison with large language models, we developed a prompt and applied GPT-4 for symptom extraction with assertion detection on a 20-note subset of the WCM internal validation notes. Furthermore, we analyzed 47,654 clinical notes from 11 health systems to conduct a population-level prevalence study, aiming to improve the understanding of PASC-symptom mentioning patterns and refine disease characterization. Leveraging the RECOVER\cite{recover2024-tc} datasets, our approach addresses key limitations of prior research by incorporating a large, diverse population and validating performance across multiple sites, which helps account for variations in clinical language and documentation styles.

\section{RESULTS}\label{results}

\subsection{PASC Lexicon}\label{pasc-lexicon}

The PASC lexicon is an ontology graph consisting of 25 symptom categories, 798 finer-grained UMLS concepts, and their synonyms (\textbf{Figure \ref{fig:lexicon}}). The complete symptom lexicon is available in \textbf{Supplementary File 1}.

\begin{figure}
    \centering
    \includegraphics[width=\linewidth]{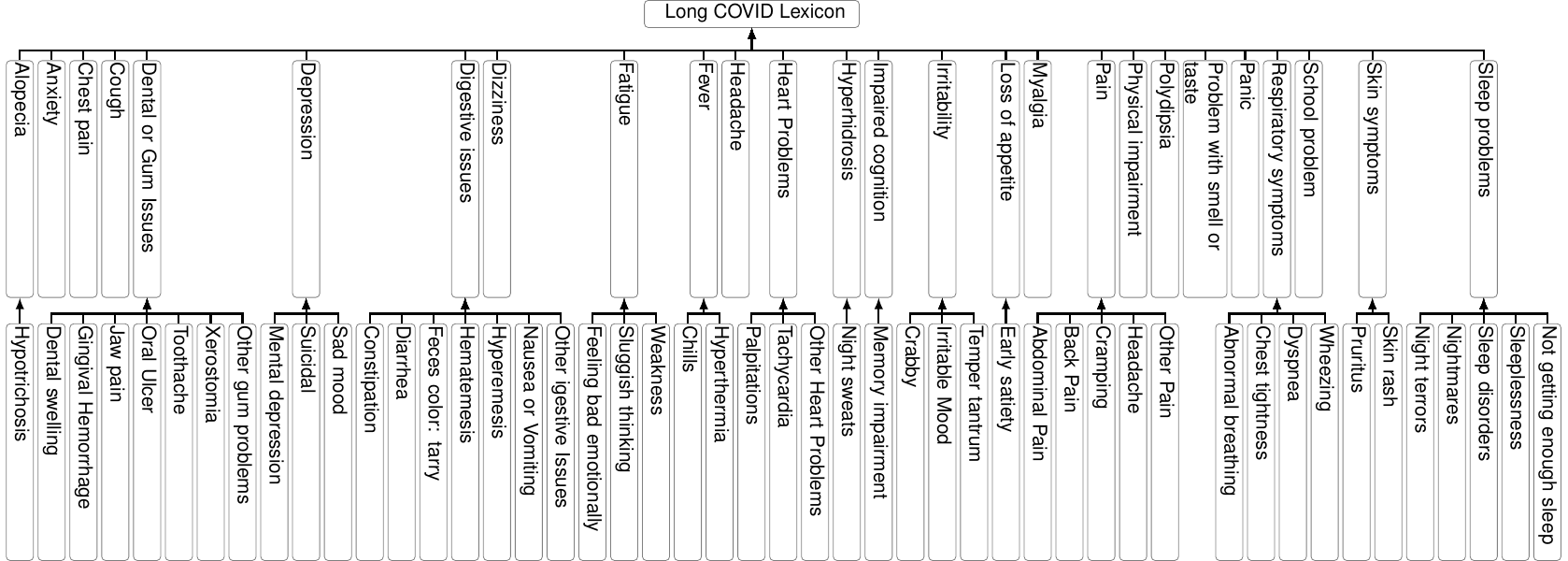}
    \caption{PASC Lexicon with Examples of Representative Symptoms.}
    \label{fig:lexicon}
\end{figure}

The 25 categories, together with the initial symptom lexicon, were identified by physicians (led by CF). The categories and lexicon were then manually mapped and consolidated using UMLS and SNOMED CT. To make the lexicon comprehensive, we included all synonyms from the UMLS concepts and the lexicon extracted from PASCLex\cite{Wang2022-ua}. We further leveraged the Broader-Narrower relationship (hierarchies between concepts) and included all children of a concept in the initial lexicon.

\subsection{A Hybrid NLP Pipeline for PASC Symptom Extraction}\label{a-hybrid-nlp-pipeline-for-pasc-symptom-extraction}

\begin{figure}
    \centering
    \includegraphics[width=.7\linewidth]{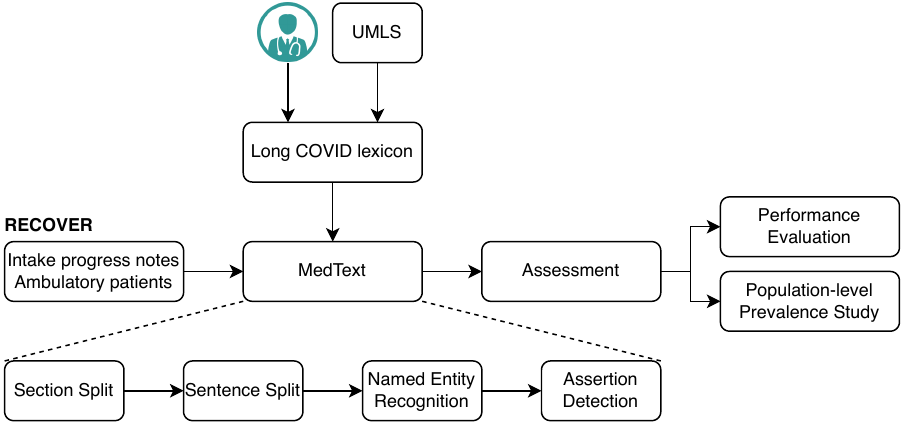}
    \caption{The architecture of the NLP pipeline.}
    \label{fig:pipeline}
\end{figure}

We developed MedText, a hybrid NLP pipeline that integrates both rule-based modules and deep learning models at different stages (\textbf{Figure \ref{fig:pipeline}}). At a high level, MedText employs a text preprocessing module to split the clinical notes into sections and sentences. It then uses the \textbf{PASC lexicon} and a rule-based NER module to extract PASC symptoms from clinical notes. Finally, a BERT-based assertion detection module is employed to determine whether the extracted symptom is positive or not (e.g., ``there is no diarrhea'').

\subsection{Quantitative Results for Assertion Detection}\label{quantitative-results-for-assertion-detection}

In this study, we used three popular domain-specific pretrained BERT models, BioBERT, ClinicalBERT, and BiomedBERT\cite{Lee2020-al, Huang2019-nf, Gu2020-oj}, to develop the assertion detection module in MedText. All BERT models were fine-tuned using the combination of the WCM Training Set and public i2b2 2010 assertion dataset\cite{Uzuner2011-gz} (See Methods).

We evaluated the performance of assertion detection using the clinical notes in the WCM Internal Validation and Multi-site External Validation sets. Performance was measured regarding precision, recall, and F1-score (the harmonic mean of the precision and recall) across all the symptoms in these two datasets.

\begin{figure}
    \centering
    \includegraphics[width=\linewidth]{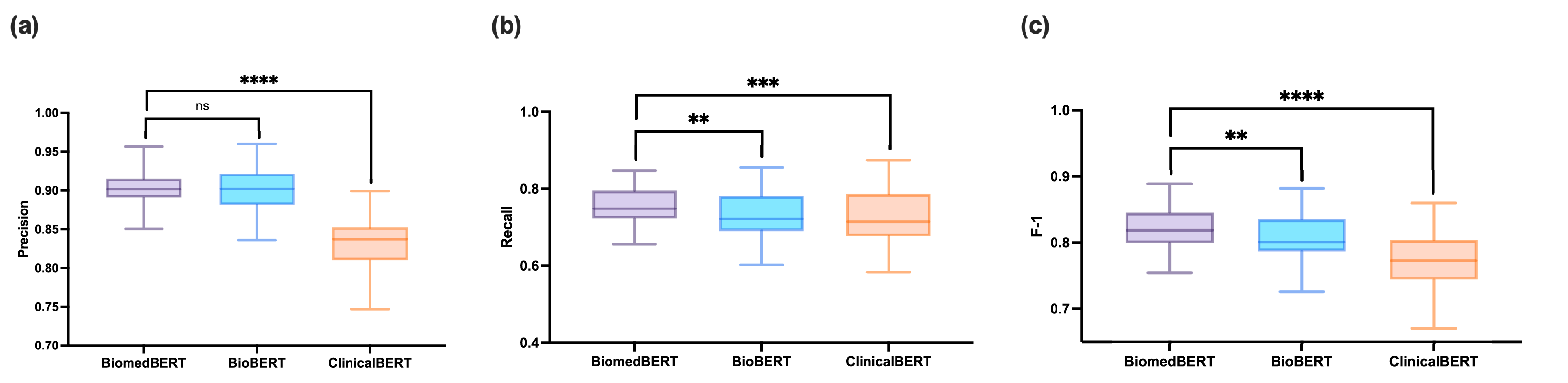}
    \caption{The performance of three BERT variants on the WCM internal validation set. ns - not significant; ** - $P\leq 0.01$; *** - $P\leq0.001$; **** - $P\leq0.0001$.}
    \label{fig:bert}
\end{figure}

\textbf{Figure \ref{fig:bert}} presents a comparative performance analysis on the WCM Internal Validation dataset. Among the three models, the BiomedBERT consistently shows superior performance across all metrics, particularly in an average recall of $0.75\pm 0.040$ (95\% CI: 0.676--0.834) and an average F1 score of $0.82\pm 0.028$ (95\% CI: 0.766 -- 0.876).

\begin{figure}
    \centering
    \includegraphics[width=\linewidth]{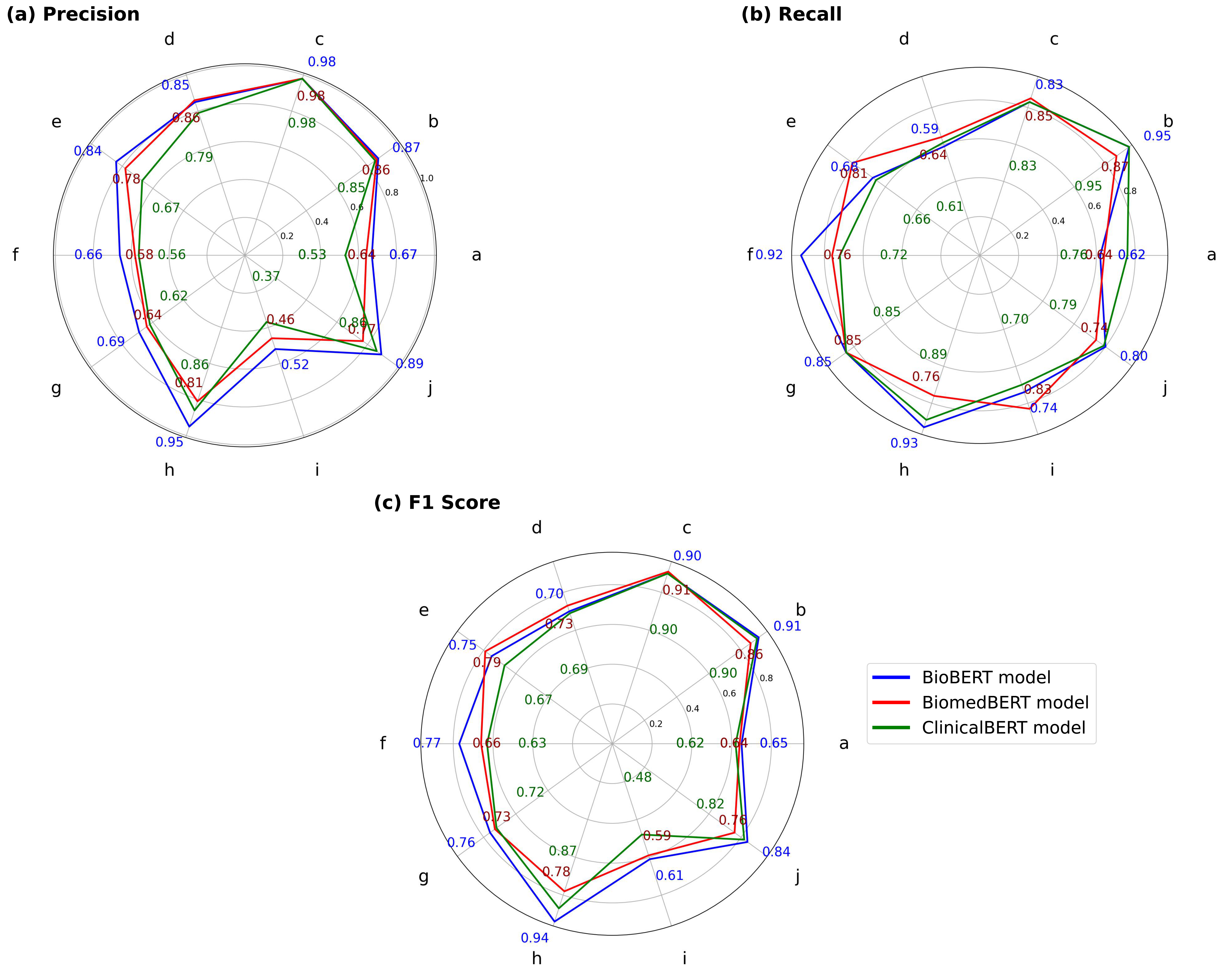}
    \caption{Multi-site External Validation. Radar Chart for the performance metrics---(a) precision, (b) recall, and (c) F1-score---of the three fine-tuned MedText-BERT pipeline variants on the 100-note non-WCM multi-site external validation set. The metrics are computed for the positive (i.e., ``Present'') symptom mentions. The BERT-based models trained/fine-tuned in different scenarios for assertion detection are: BioBERT fine-tuned, BiomedBERT fine-tuned, BiomedBERT benchmark, and ClinicalBERT fine-tuned from left to right in each subfigure.}
    \label{fig:multisite}
\end{figure}

\textbf{Figure \ref{fig:multisite}} presents a comparative performance analysis on the Multi-site External Validation dataset. The precision, recall, and F1 scores show no significant differences across the models (i.e., $P>0.05$ in t-tests). BiomedBERT achieved an average recall of 0.775 (95\% CI: 0.726 -- 0.825) and an average F1 score of 0.745 {[}95\% CI: 0.697 -- 0.796{]}) and ClinicalBERT (recall of 0.775 {[}95\% CI: 0.713 -- 0.838{]}) and F1 of 0.730 {[}95\% CI: 0.649 -- 0.812{]}), lower than BioBERT (recall of 0.792 {[}95\% CI: 0.715 -- 0.868{]} and F1 of 0.782 {[}95\% CI: 0.716 -- 0.848{]}). The average precision of BiomedBERT (0.737 {[}95\% CI: 0.646 -- 0.829{]}) is not significantly different from BioBERT (0.792 {[}95\% CI: 0.704 -- 0.880{]}), but is higher than ClinicalBERT (0.710 {[}95\% CI: 0.598 -- 0.822{]}).

\subsection{Frequency Analysis of PASC Symptoms in The Population-Level Prevalence Study}\label{frequency-analysis-of-pasc-symptoms-in-the-population-level-prevalence-study}

Based on the WCM Internal Validation and Multi-site External Validation results, the BiomedBERT model was selected for a population-level prevalence study.

\begin{figure}
    \centering
    \includegraphics[width=\linewidth]{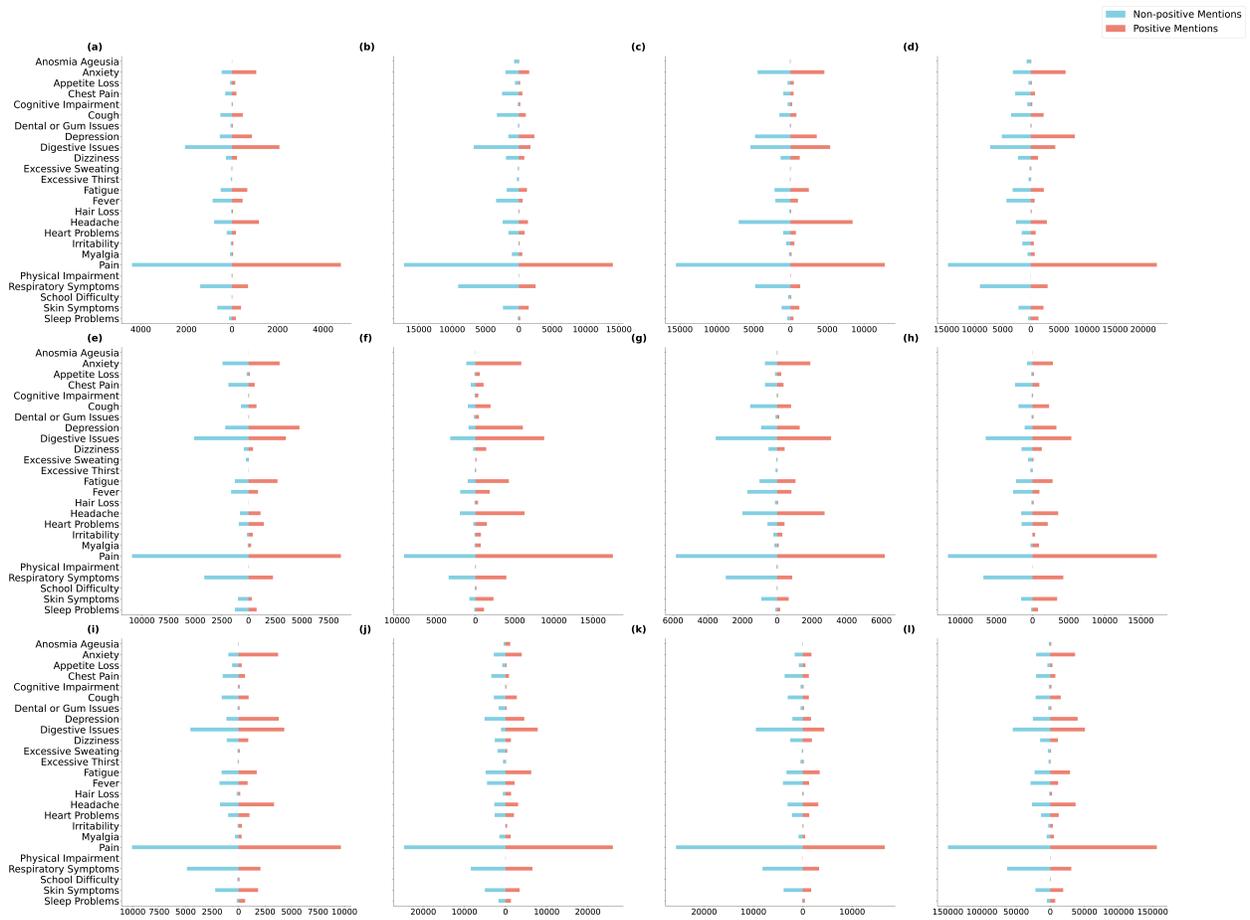}
    \caption{Frequency analysis of positive (``present'') in red and negative (``non-present'') in blue symptom category occurrences in different sites.}
    \label{fig:population}
\end{figure}

\textbf{Figure \ref{fig:population}} summarizes the positive (i.e., ``positive'') mentions and negative (i.e., ``non-positive'') mentions of 25 PASC categories identified by MedText on the Population-level Prevalence Study dataset. Among all 11 sites, the Positive mention of ``pain'' is the leading symptom mentioned. Other leading symptom categories include headache, digestive issues, depression, anxiety, respiratory symptoms, and fatigue, in potentially varying orders across different sites.

\begin{figure}
    \centering
    \includegraphics[width=\linewidth]{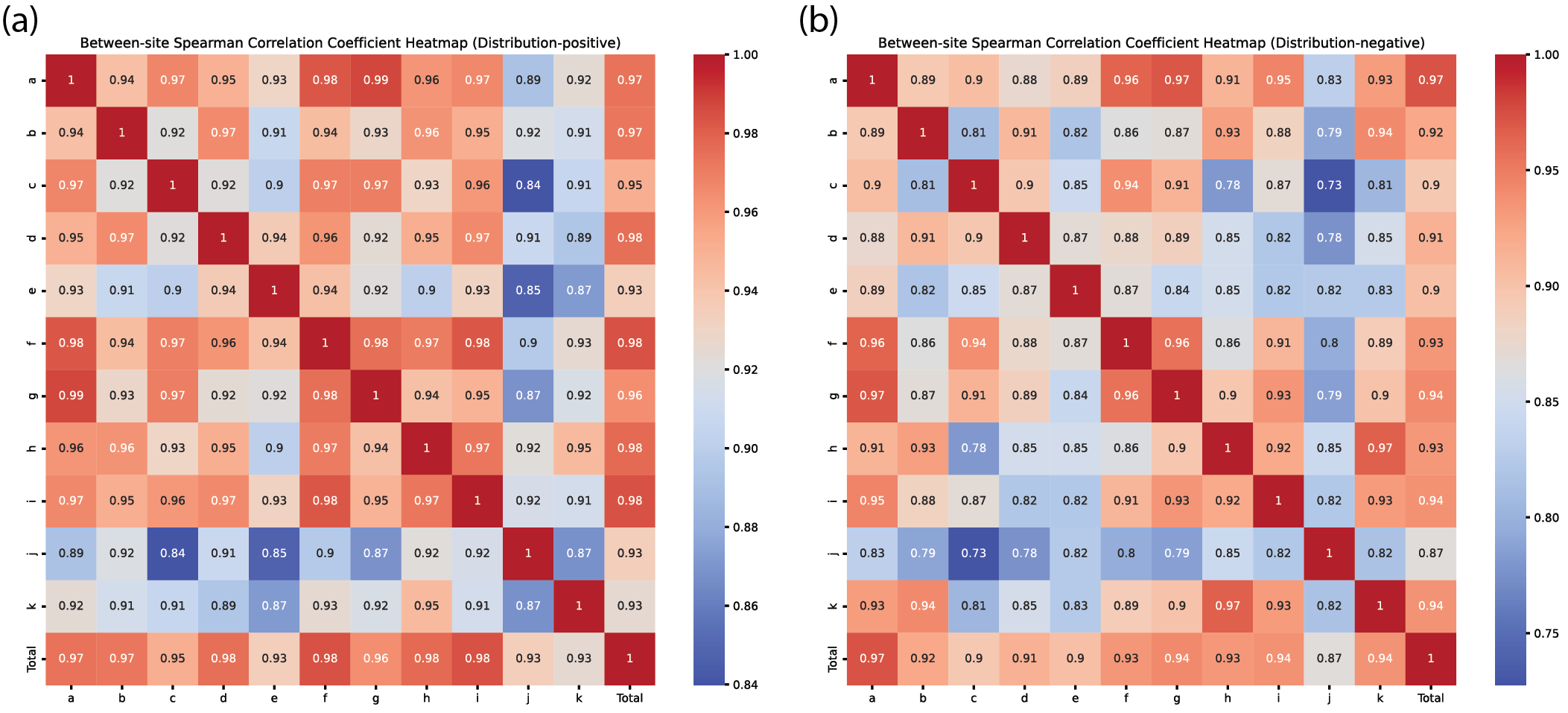}
    \caption{Population-level Prevalence Study. (A) Spearman correlation coefficients between the positive (i.e., ``present'') symptom mentioning patterns of sites and the overall dataset. (B) Spearman correlation coefficients between the negative (i.e., ``non-present'') symptom-mentioning patterns of sites and the overall dataset.}
    \label{fig:population2}
\end{figure}

\textbf{Figures \ref{fig:population2}A} and \textbf{\ref{fig:population2}B} further compare the symptom-mentioning patterns in terms of the relative frequency of the symptom categories in different sites through the Spearman correlation test. In particular, the lowest Spearman correlation coefficient for the positive symptom-mentioning patterns between any two sites and the entire dataset exceeds 0.83, while for negative symptom-mentioning patterns, it is above 0.72. In the Spearman correlation test, any pair of sites or between any site and the overall dataset showed $P<0.0001$.

\begin{figure}
    \centering
    \includegraphics[width=\linewidth]{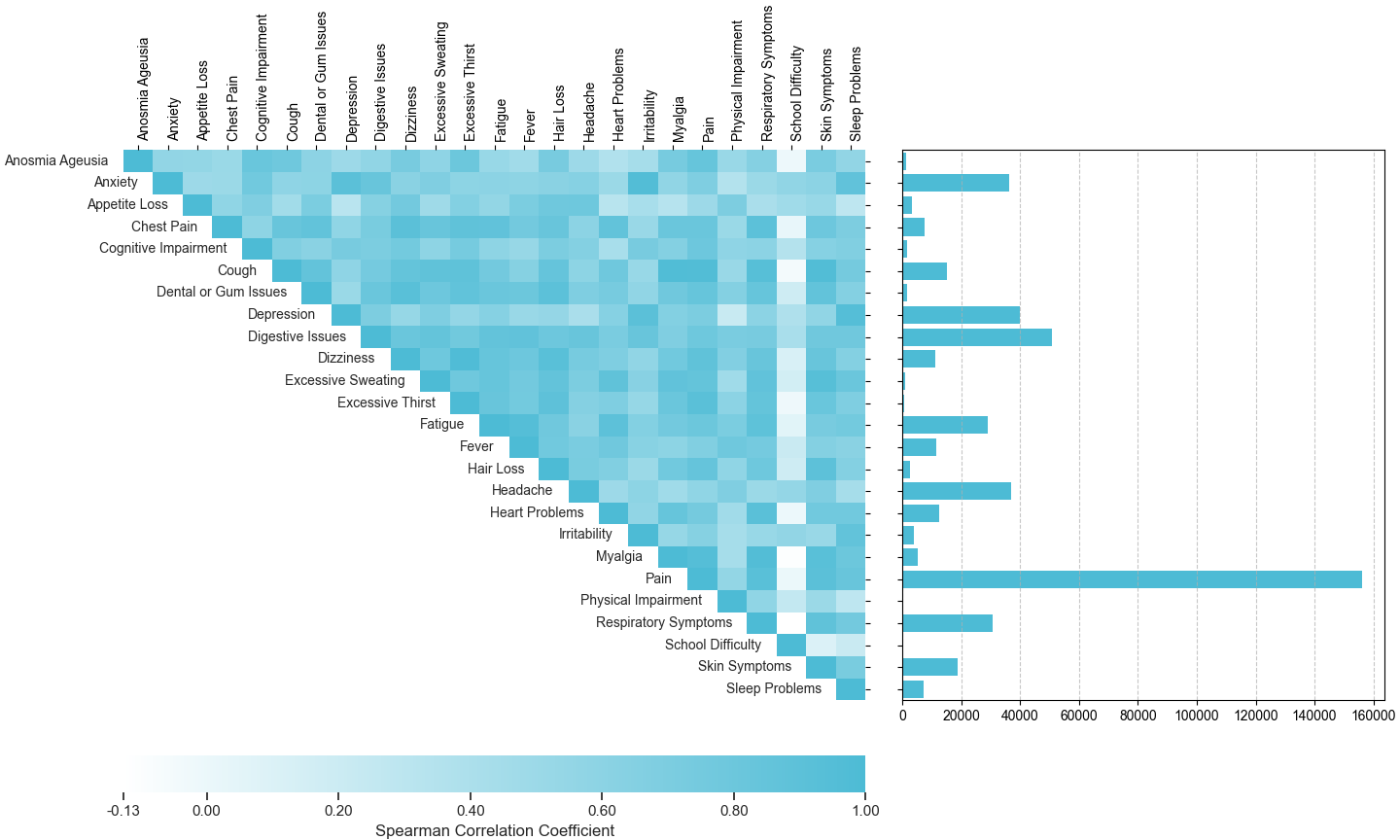}
    \caption{Cross-symptom-category correlation test and symptom-category distribution. Spearman correlation coefficients between the positive (i.e., ``Present'') symptom-mentioning patterns of symptom categories. The total count of positive symptom mentions for each symptom category is to the right of the correlation diagram.}
    \label{fig:Cross-symptom-category}
\end{figure}

\textbf{Figure \ref{fig:Cross-symptom-category}} shows the Spearman correlation coefficients for the positive mentions between each pair of the 25 symptom categories and the total number of symptom mentions.

\subsection{Processing Time}\label{processing-time}

\begin{figure}
    \centering
    \includegraphics[width=0.5\linewidth]{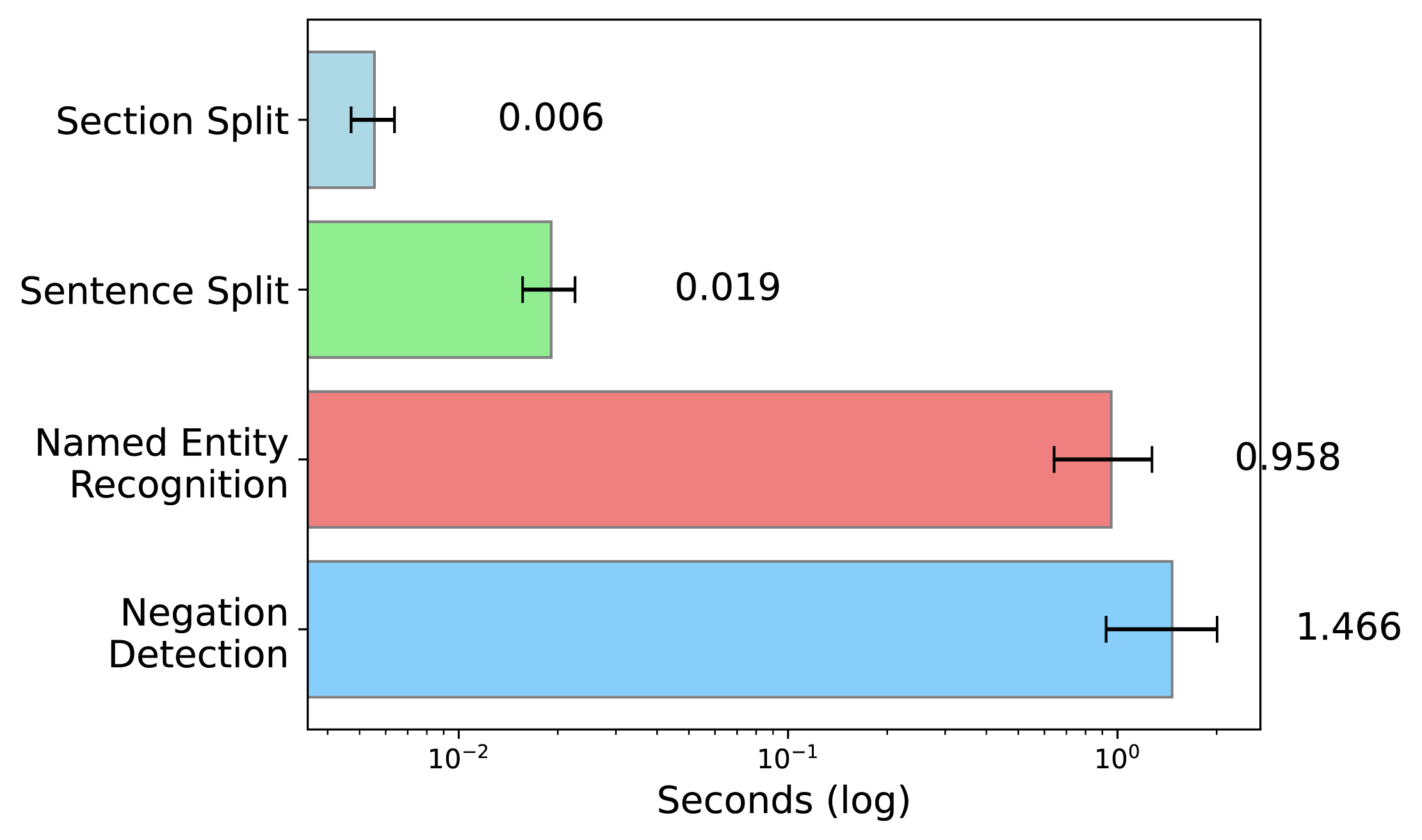}
    \caption{Module-wise runtime summary of MedText processing. The mean (shown by the values on each bar) and standard deviation of the runtime of each module in MedText were computed for the mean runtime per note across the 11 sites.}
    \label{fig:runtime}
\end{figure}

We deployed MedText on an AWS Sagemaker platform with a Tesla T4 GPU configuration, achieving efficient processing of the large-scale clinical notes datasets. Overall, MedText processed each note at an average of $2.448 \pm 0.812$ seconds across the 11 sites. \textbf{Figure \ref{fig:runtime}} shows the module-wise summary of the mean and standard deviation of runtime in seconds in the population-level prevalence study across 11 sites.

\subsection{Comparison between Rule-Based Module and GPT-4 for Symptom Extraction}\label{comparison-between-rule-based-module-and-gpt-4-for-symptom-extraction}

To strengthen our evaluation, we conducted an experiment comparing our rule-based NER module with a large language model, GPT-4, for PASC-related symptom extraction (see Methods). Using a random sample of 20 intake notes from the WCM internal validation set, we applied GPT-4 (API version: \emph{2024-05-01-preview}) with a prompt specifying the 25 predefined symptom categories. GPT-4 generated 189 symptom mentions, which ZB manually reviewed to assess correctness---specifically, whether each mention existed in the note and corresponded to a valid symptom or synonym. This review identified 9 incorrect mentions, yielding a precision of 95.24\% for GPT-4-based NER.

To estimate recall of the NER module in our hybrid NLP pipeline, we constructed a proxy ground truth by taking the union of all verified correct mentions identified by GPT-4 (180) and our rule-based NER module (640) on these 20 notes. This union set contains 706 verified correct mentions. Based on this reference set, the rule-based module achieved a recall of 90.65\%.

In addition, our prompt also required GPT-4 to generate assertion labels. These assertion detection results produced by GPT-4 were manually reviewed for correctness, which achieved a weighted F1 score of 97.78\% and a balanced accuracy of 97.72\% on the 180 correctly recognized symptom mentions.

\textbf{DISCUSSION}

In this study, we developed a hybrid NLP pipeline combining rule-based and deep learning techniques, leveraging state-of-the-art techniques\cite{medtext-nb, Wang2022-wh} for clinical text preprocessing and symptom extraction. Our pipeline not only identified relevant symptoms of PASC but also accurately evaluated whether the symptoms are truly present or absent. This pipeline includes a PASC lexicon we developed with clinical specialists, with 25 symptom categories and 798 finer-grained UMLS concepts and synonyms. We adapted a BERT-based module for symptom assertion detection. For model training and evaluation, we curated 160 progress notes from 11 health systems as part of the RECOVER initiative network across the U.S. Experiments on the WCM Internal Validation set and the Multi-site External Validation set show that BiomedBERT achieved the overall best performance in assertion detection. Our model showed high precision and recall, achieving an average F1 score of 0.82 in internal validation at one site and 0.76 in external validation across 10 sites for assertion detection, showing good generalizability across sites. Additionally, we collected 47,654 progress notes for a population-level prevalence study on PASC symptom-mention patterns. Our pipeline processed each note at an average speed of 2.448±0.812 seconds. Spearman correlation tests showed $\rho > 0.83$ for positive mentions and $\rho > 0.72$ for negative mentions, both with $P < 0.0001$. These results demonstrate that our pipeline effectively, consistently, and efficiently captures PASC symptoms across multiple health systems. Our population-level prevalence study provided novel insights and may advance the understanding of the symptom-mentioning patterns related to PASC, shedding light on a more comprehensive understanding of PASC for future research and clinical practice.

The hybrid NLP pipeline we developed is tailored explicitly to PASC, while previous research has explored assertion detection in different clinical contexts, e.g., the BERT-based\cite{vanAken2021assertion} and prompt-based\cite{Wang2022-po}. In particular, BERT-based models applied to Chia\cite{kury2020chia} achieved an F1 score of $\sim 0.77$ for ``Present'' and a micro-averaged F1 score of $\sim 0.72$. Although these models claimed over 0.9 in both the F-1 score for ``Present'' and the micro-averaged F1 score evaluated on conventional benchmarking datasets, i2b2 2010\cite{Uzuner2011-gz}, BioScope\cite{vincze2008bioscope}, MIMIC-III\cite{johnson2016mimic-iii}, and NegEx\cite{harkema2009context}, none of them have addressed the unique challenges posed by PASC-related narratives. Our model's performance on the curated RECOVER datasets achieved an average F1 score of 0.82 in internal validation and 0.76 in 10-site external validation via BiomedBERT. This demonstrates that domain-specific models remain necessary compared to general-domain language models, given the nascent and rapidly evolving nature of PASC-related medical narratives. The lack of standardized terminology\cite{Wang2022-ua}, coupled with the limited availability of annotated training data\cite{wen2024case} and the heterogeneous symptomatology of PASC\cite{fritsche2023characterizing}, presents different challenges that are not as prevalent in more established clinical note datasets for NLP tasks. Comparative analyses of holistic processing speeds across different NLP pipelines with BERT-based modules for unstructured clinical notes are not extensively documented in the literature, and the differences between datasets and experimental setups can undermine fair comparison. Nonetheless, our model offers a competitive runtime of $\sim 2.5$ seconds per note compared to LESA-BERT, a model originally designed for patient message triage, which required approximately 212.6 seconds to process its test set on a CPU with a batch size of 1. Its distilled versions, such as Distil-LESA-BERT-6 and Distil-LESA-BERT-3, showed inference times of 79.8 seconds and 40.8 seconds, respectively\cite{si2020students}. This demonstrates the efficiency of our pipeline.

We conducted error analysis during multi-site external validation and identified site-specific performance limitations, revealing potential causes that may affect model effectiveness. First, inconsistent manual annotations, e.g., ``as needed'' cases, led to varied interpretations of symptoms as rigorously ``positive'', ``hypothetical'' in EHR was identified as ``non-positive''; while phrases containing ``history of'', symptoms were often misclassified as ``positive''. Second, sentences listing multiple symptoms after the negation phrase ``negative for'' posed challenges. Excessive spacing in these instances caused the model to misinterpret the context, leading to symptoms frequently misclassified as ``positive'', likely due to their distance from the negation keyword. Third, symptoms that were excluded from the performance evaluation, e.g., due to unresolved mismatches, could affect the NLP pipeline's overall performance. Hopefully, these findings may provide clues for those working toward enhancing their model's ability to address more intricate textual contexts or classify temporal context.

To date, we have not found any study reported applying Large Language Models (LLMs) (e.g., general-purpose models ChatGPT\cite{chatgpt-oz}, Gemini\cite{Hirosawa2024-np, gemini}, and open-source, medical domain-specific models OpenBioLLM-70B\cite{llama3-de} and Llama-3-8B-UltraMedical\cite{llamaUltraMedical-oy}) on PASC-related symptom extraction with assertion detection. To strengthen our evaluation, we randomly selected 20 notes from the 30-note WCM internal validation subset and applied GPT-4 to perform both NER and assertion detection, and compared with the rule-based NER module in symptom extraction. Without specifying PASC-specific lexicon, GPT-4 was able to achieve better performance in assertion detection on the 180 manually verified correctly identified symptoms, with a weighted F1 score of 97.78\%. However, it also missed a notable amount of symptom/synonym-related tokens compared to the 640 correct symptom mentions by our rule-based NER module on the same 20 notes.

Our study has limitations: (i) Our current performance evaluation focused on the positive or negative mentions of symptoms already identified in the MedText pipeline. However, whether all possible symptoms are captured remains unclear. (ii) We excluded 18 symptom mentions from the performance evaluation due to unresolved mismatches that may affect the results of symptom extraction. (iii) Our primary goal in this work is to develop a generalizable NLP pipeline for extracting PASC-relevant symptoms from clinical narratives, with intake notes serving as an initial use case. While the intake notes are often the earliest comprehensive documentation of a patient's initial concerns, they may overrepresent acute or prominent symptoms, potentially biasing the extracted distributions. To address these limitations, we will enhance our pipeline in the future via: (i) expanding the annotated datasets in collaboration with clinicians to annotate all possible symptoms given a note, (ii) integrating structured EHR data and COVID-related symptoms extracted from unstructured clinical notes, (iii) evaluating how different note types contribute to identifying high-risk patients and explore how the extracted outputs could be integrated into clinical workflows, for example, by flagging patients for specialist referral or follow-up care. Since our pipeline itself is not restricted to intake notes, it can be directly applied to other types of notes.

All the above demonstrate that our hybrid NLP pipeline, namely MedText, can extract symptoms effectively, efficiently, and robustly. Our MedText may contribute to (i) unlocking rich clinical information from narrative clinical notes, (ii) assisting clinicians in diagnosing PASC efficiently and precisely, (iii) supporting predictive modeling and risk assessment about PASC, (iv) facilitating the PASC clinical decision support system, and (v) adapting our pipeline for exact symptoms of other diseases may enable large-scale research and public health insights. However, MedText may not capture all nuanced clinical details, highlighting the importance of clinical review to ensure accuracy and completeness in symptom identification.

\section{METHODS}\label{methods}

\subsection{Study cohort}\label{study-cohort}

We collected 47,814 clinical notes from 11 sites within the RECOVER network: Weill Cornell Medicine (WCM), Medical College of Wisconsin, Cincinnati Children's Hospital Medical Center, The Children's Hospital of Philadelphia, University of Missouri, Nationwide Children's Hospital, Nemours Children's Health System, Oregon Community Health Information Network, Seattle Children's, UT Southwestern Medical Center, and Montefiore Medical Center. The data follow the Observational Medical Outcomes Partnership (OMOP) Common Data Model (CDM), with the Institutional Review Board (IRB) approval obtained under Biomedical Research Alliance of New York (BRANY) protocol \#21-08-508.

\subsection{Curating Data for PASC Symptom Review and NLP Pipeline Evaluation}\label{curating-data-for-pasc-symptom-review-and-nlp-pipeline-evaluation}

\begin{figure}
    \centering
    \includegraphics[width=0.5\linewidth]{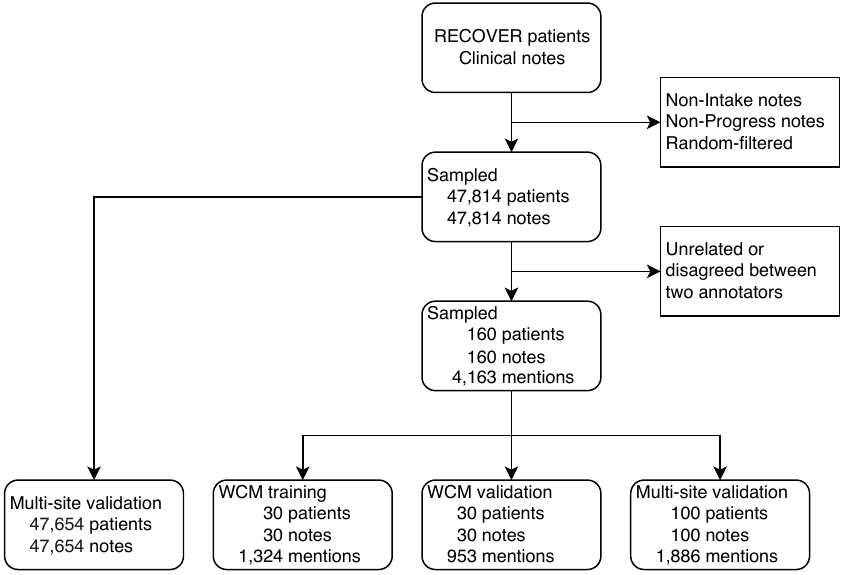}
    \caption{Data construction workflow.}
    \label{fig:workflow}
\end{figure}

For pipeline development and validation, we sampled 60 ambulatory patients from WCM and 100 from 10 other sites in the RECOVER network (10 patients/site) (\textbf{Table \ref{tab:dataset}}). Then, we extracted each patient's intake progress notes for human annotators. Here, we hypothesize that the intake notes contain the most detailed list of problems for each patient. The 60 WCM notes form the ''Model Development Dataset'', in which 30 were used as the WCM Training set, and the other 30 as the WCM Internal Validation set. The 100 notes from 10 other sites form the Multi-site External Validation set (\textbf{Figure \ref{fig:workflow}}). We also integrated the publicly available 2010 i2b2 assertion dataset\cite{Uzuner2011-gz} with the WCM training set for model fine-tuning.

\begin{table}
    \caption{Datasets curated from the 11 sites of the RECOVER Initiative.}
    \label{tab:dataset}
    \scriptsize
    \input{tables/table1}
\end{table}

\begin{figure}
    \centering
    \includegraphics[width=0.9\linewidth]{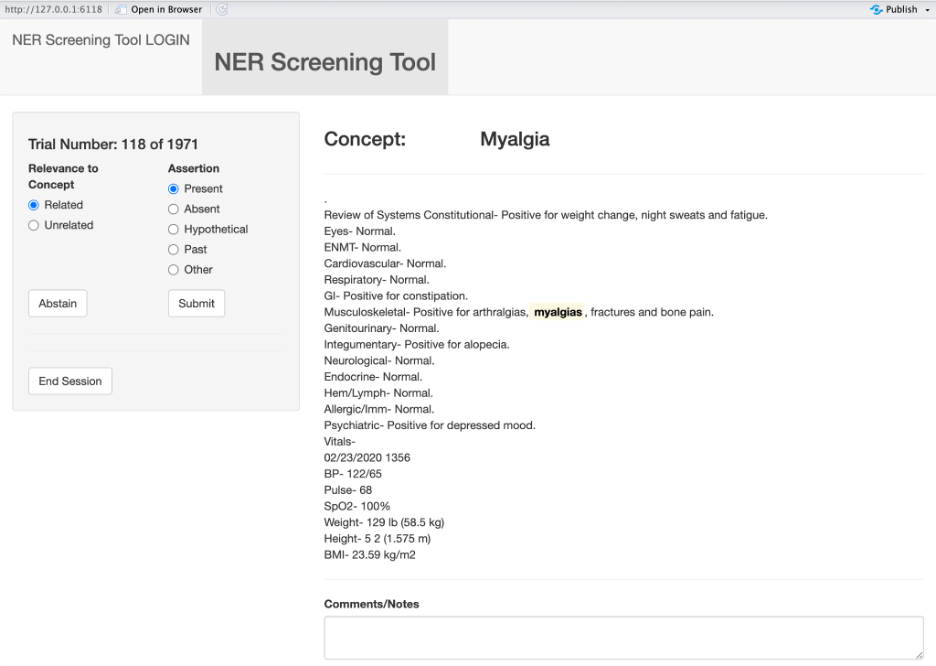}
    \caption{Example screenshot for Screen\_Tool. Screen\_Tool is an open-source R-based software for manual annotation of symptom mention.}
    \label{fig:screenshot}
\end{figure}

For annotation, we designed a qualified review pipeline to generate manual annotations for symptom mentions. We first applied MedText to extract symptoms from the notes. Then, we used Screen-Tool, an open-source R software developed by TB, to determine the assertion status for each symptom mention (e.g., \textbf{Figure \ref{fig:screenshot}}). Screen-Tool displays each extracted symptom highlighted in its context (the passage it belongs to), along with the symptom category identified by MedText as its concept. The annotators are required to determine whether this token is ``related'' or ``unrelated'' to this concept, as well as which of the five statuses it belongs to: ``present'', ``absent'', ``hypothetical'', ``past'', or ``other''. For performance validation of assertion detection, these statuses are eventually mapped into a binary status: positive vs. non-positive. Specifically, ``present'' mentions are mapped to positive, while all the other statuses of mentions are deemed Negative.

Two annotators (ZB and ZX) independently reviewed and annotated each mention of extracted symptoms. We used Cohen's Kappa (CK) metric to measure inter-rater agreement (IRA). This process achieved a value of 0.98 on the WCM annotated subset (i.e., including both the WCM training and WCM internal validation sets), and 0.99 on the multi-site external validation set. All mentions that had disagreements between the two annotators or were unrelated were removed from the dataset. Specifically, 24 out of 2,301 mentions ($<1.05\%$) were removed from the WCM annotated subset, and 18 out of 1,886 ($< 0.5\%$) were removed from the multi-site external validation set. Among the 4,187 mentions extracted from 160 annotated notes, 12 were `unrelated' NER results where the trigger token did not match a recognized symptom or synonym. This yields an overall NER precision of 99.7\%, with 99.5\% (11 unrelated) on the WCM subset and 99.9\% (1 unrelated) on the multi-site external validation set.

\subsection{Constructing PASC Lexicon}\label{constructing-pasc-lexicon}

We first compiled an initial symptom lexicon with 25 categories for PASC-related terms, based on input from subject matter experts (SMEs) and a literature review\cite{Michelen2021-bp}. PASC symptoms were defined as patient characteristics occurring in the post-acute COVID-19 period. Next, we searched the symptoms in the UMLS Metathesaurus® (version 2023AB) in the English language, with vocabulary sources of ``SNOMEDCT\_US'' and ``MeSH''. We included concepts that can be strictly matched in UMLS, together with their children (narrow concepts). To make the lexicon comprehensive, we included all synonyms from the UMLS concepts and the lexicon extracted from PASCLex\cite{Wang2022-ua}. This process yielded a hierarchical knowledge graph comprising terms and keywords that define 798 symptom sub-concepts and synonyms, organized into 25 categories (\textbf{Supplementary File 2}).

\subsection{Developing a Hybrid NLP Pipeline for PASC Symptom Extraction}\label{developing-a-hybrid-nlp-pipeline-for-pasc-symptom-extraction}

We developed a hybrid NLP pipeline that combines rule-based and deep-learning modules to mine PASC symptoms. At a high level, MedText employs a text preprocessing module to split the clinical notes into sections and sentences. It then uses a rule-based NER module that utilizes a clinician-curated PASC lexicon to extract PASC symptoms from clinical notes. It then employs a BERT-based assertion detection module to assess if the extracted symptom is absent (e.g., ``there is no diarrhea'') (\textbf{Figure \ref{fig:workflow}}).

MedText\cite{medtext-nb} is an open-source clinical text analysis system developed with Python. It offers an easy-to-use text analysis pipeline, including de-identification, section segmentation, sentence split, NER, constituency parsing, dependency parsing, and assertion detection. For this study, we utilized modules ranging from section segmentation and sentence split to NER and assertion detection (\textbf{Figure \ref{fig:workflow}}).

The section split module divides the report into sections. This rule-based module uses a list of section titles to divide the notes. We used rules from medspacy, which were adapted from SecTag and expanded through practice\cite{denny2009evaluation}. The sentence split module splits the report into sentences using NLTK\cite{bird2006nltk}. The NER module recognizes mention spans of a particular entity type (e.g., PASC symptoms) from the reports. Technically, it uses a rule-based method via SpaCy's PhraseMatcher to efficiently match large terminology lists. Here, the NER step identifies and clarifies concepts of interest according to the 798 symptoms and their synonyms in the PASC lexicon. The assertion detection module determines the status of the recognized concept based on its context within the sentence\cite{Wang2022-po}, as detailed below in the assertion detection module development.

Additionally, MedText features a flexible modular design, provides a hybrid text processing schema, and supports raw text processing and local processing. It also adopts BioC\cite{comeau2013bioc} as the unified interface, and standardizes the input/output into a structured representation compatible with the OMOP CDM. This allows us to deploy MedText and validate its performance across multiple, disparate data sources. In this study, we deployed MedText on an AWS Sagemaker platform with a Tesla T4 GPU configuration and 15 GiB of CPU memory.

\subsection{Developing the Assertion Detection Module}\label{developing-the-assertion-detection-module}

In this study, we used three popular domain-specific pretrained BERT models, BioBERT, ClinicalBERT, and BiomedBERT\cite{Huang2019-nf, Gu2020-oj, Uzuner2011-gz}, to develop classifiers for the assertion detection step in the MedText pipeline after extracting symptom mentions. Detailed descriptions of these pretrained BERT models and their hyperparameter settings are provided in the \textbf{Table \ref{tab:hyperparameters}}.

\begin{table}[]
    \caption{Hyperparameters for BERT-based assertion detection model training.}
    \label{tab:hyperparameters}
    \input{tables/table2}
\end{table}

To construct training sets for model development, we utilized a publicly available 2010 i2b2 assertion dataset of clinical notes\cite{medtext-nb, Uzuner2011-gz}, which consists of 4,359 sentences and 7,073 labels collected from three sites: Partners Healthcare, Beth Israel Deaconess Medical Center, and the University of Pittsburgh Medical Center\cite{medtext-nb, Uzuner2011-gz}, and the WCM Model Training set to construct training sets for model development. In particular, 3,055 sentences and 4,243 mentions from the public i2b2 2010 assertion dataset\cite{medtext-nb, Uzuner2011-gz} merged with the 30-note WCM Model Training set were used as the training set for fine-tuning the BERT-based assertion detection models. This integrated dataset is referred to as the i2b2\&WCM-merged training set. To evaluate the effectiveness of our fine-tuning approach, we conducted comparative analyses among BiomedBERT\cite{Gu2020-oj}, BioBERT\cite{Lee2020-al}, and ClinicalBERT\cite{Huang2019-nf} - all of which were trained on the i2b2\&WCM-merged training set. We used the WCM internal validation set (30 notes, 350 sentences, and 953 mentions) for model selection and the 10-site multi-site external validation set (100 notes, 1,113 sentences, and 1,886 mentions) for performance evaluation (\textbf{Table 1}). This direct comparison offered insights into the relative strengths of each model in handling biomedical text. The multi-category prediction task of multiple possible assertion statuses from the original BERT models was revised into a binary prediction task. Specifically, only the ``present'' status is mapped to present with a label of 1. In contrast, all other possible statuses, including ``absent'', ``hypothetical'', and any form of ``uncertain'', are mapped to negative with a label of 0.

\subsection{Evaluation Plan}\label{evaluation-plan}

We explored three different models of assertion detection in the MedText pipeline and considered two different application scenarios: the \textbf{numerical performance evaluation} using the smaller datasets with manually annotated ground-truth labels (i.e., the Internal Validation set, and the Multi-site External Validation set) and the \textbf{population-level prevalence study} using approximately 5,000 clinical notes from each site from the 11 sites of the RECOVER network (i.e., the population statistics set). The patient demographics (race, ethnicity, age, and sex) across the 11 RECOVER sites in the population-level prevalence study are summarized in \textbf{Table \ref{tab:summary}}.

\begin{table}
    \caption{Summary of patient demographics. Patient demographics (race, ethnicity, age, and sex) across the 11 RECOVER sites.}
    \label{tab:summary}
\resizebox{\textwidth}{!}{
    \input{tables/table3}
}
\end{table}

In the numerical performance evaluation, each model was assessed using precision, recall, and F1 score. For each trained and fine-tuned model, bootstrapping was applied in 100 iterations on the WCM Internal Validation set. The performance of these models was evaluated using the same metrics, along with their mean values, standard deviation, and 95\% confidence intervals.

On the prevalence of PASC-related symptoms at the population level, we analyzed symptom mentions identified by the optimal configuration of our NLP pipeline across 11 RECOVER sites. Using the MedText pipeline, we analyzed 47,654 notes from these sites as a \textbf{Population-level Prevalence Study} dataset (\textbf{Figure \ref{fig:population}}). Specifically, this dataset included 5,000 intake progress notes from unique patients per site, except for three sites: Nationwide had 3,680 notes, Nemours had 2,132, and Seattle had 1,842. We then applied the Spearman rank correlation test to compare symptom-mentioning patterns across sites. We conducted pairwise Spearman tests between the sites and symptom categories, for positive and negative mentions, respectively. For between-site analysis, the symptom mentions over the 25 symptom categories of each site form the symptom-mentioning vector for this site. For between-site analysis, the symptom-mentions across the 11 sites of each symptom category form the symptom-mentioning vector of this symptom category.

\subsection{Symptom Extraction and Assertion Detection Using Large Language Model: GPT-4}\label{symptom-extraction-and-assertion-detection-using-large-language-model-gpt-4}

To compare our hybrid NLP pipeline with large language models for PASC symptom extraction and assertion detection, we conducted an experiment using GPT-4 (API version: \emph{2024-05-01-preview}). We randomly selected 20 intake notes from the subset used for WCM internal validation. GPT-4 was prompted with only the 25 predefined PASC symptom categories and instructed to extract all symptom mentions along with their assertion status (i.e., positive or negative). The model was allowed to rely on its internal knowledge to recognize synonymous expressions without being provided explicit lexicons. The specific prompt is provided in \textbf{Supplementary File 3}.

Each output mention was manually reviewed (by ZB) to determine whether the identified phrase existed in the original note and correctly matched a symptom or its synonym. To create a proxy ground truth for recall estimation, we took the union of all verified correct mentions extracted by either GPT-4 or the rule-based NER pipeline. Using this set, we evaluated recall for the rule-based method and precision for GPT-4. The assertion labels assigned by GPT-4 were also manually reviewed for correctness.

\section*{ACKNOWLEDGMENTS}

\textbf{Funding}: This study is part of the NIH Researching COVID to Enhance Recovery (RECOVER) Initiative, which seeks to understand, treat, and prevent the post-acute sequelae of SARS-CoV-2 infection (PASC). This research was funded by the National Institutes of Health (NIH) Agreement OTA OT2HL161847 as part of the Researching COVID to Enhance Recovery (RECOVER) research program.

\textbf{Ethics Oversight}: Institute Review Board (IRB) approval was obtained under Biomedical Research Alliance of New York (BRANY) protocol \#21-08-508. As part of the Biomedical Research Alliance of New York (BRANY IRB) process, the protocol has been reviewed in accordance with the institutional guidelines. The Biomedical Research Alliance of New York (BRANY) waived the need for consent and HIPAA authorization. IRB oversight was provided by the Biomedical Research Alliance of New York, protocol \# 21-08-508-380.

\textbf{Data Availability Statement:} Data utilized for this study were obtained from the PCORnet-RECOVER Amazon Warehouse Services (AWS) enclave, which comprises 40 participating sites from PCORnet\footnote{\url{https://pcornet.org/}}. Please send all data questions or access requests to the corresponding author, who will direct them accordingly.

We acknowledge the following RECOVER-EHR Consortium Members:

PCORnet Core Contributors

\textbf{Louisiana Public Health Institute}: \emph{Thomas W. Carton, PI}, Anna Legrand, Elizabeth Nauman

\textbf{Weill Cornell Medicine}:\emph{Rainu Kaushal, PI}, \emph{Mark Weiner, PI}, Sajjad Abedian, Dominique Brown, Christopher Cameron, Thomas Campion, Andrea Cohen, Marietou Dione, Rosie Ferris, Wilson Jacobs, Michael Koropsak, Alexandra LaMar, Colby Lewis V., Dmitry Morozyuk, Peter Morrisey, Duncan Orlander, Jyotishman Pathak, Mahfuza Sabiha, Edward J. Schenck, Catherine Sinfield, Stephenson Strobel, Zoe Verzani, Fei Wang, Zhenxing Xu, Chengxi Zang, Yongkang Zhang

Children's Hospital of Philadelphia: \emph{L. Charles Bailey, mPI}, \emph{Christopher B. Forrest, mPI}, Rodrigo Azuero-Dajud, Andrew Samuel Boss, Morgan Botdorf, Colleen Byrne, Peter Camacho, Abigail Case, Kimberley Dickinson, Susan Hague, Jonathan Harvell, Miranda Higginbotham, Kathryn Hirabayshi, Sandra Ilunga, Rochelle Jordan, Aqsa Khan, Vitaly Lorman, Nicole Marchesani, Sahal Master, Jill McDonald, Nhat Nguyen, Hanieh Razzaghi, Qiwei Shen, Alexander Shorrock, Levon H. Utidjian, Kaleigh Wieand

PCORnet Data Contributors

Albert Einstein College of Medicine \emph{Parsa Mirhaji, PI}, Selvin Soby \textbar{} Cincinnati Children's Hospital Medical Center \emph{Nathan M. Pajor, PI}, Jyothi Priya Alekapatti Nandagopal \textbar{} Children's Hospital of Philadelphia \emph{L. Charles Bailey, mPI}, \emph{Christopher B. Forrest, mPI} \textbar{} Medical College of Wisconsin \emph{Reza Shaker, PI}, \emph{Bradley W. Taylor, PI}, Alex Stoddard \textbar{} Nationwide Children's Hospital \emph{Kelly Kelleher, PI} \textbar{} Nemours/Alfred I. duPont Hospital for Children \emph{H. Timothy Bunnell, PI}, Daniel Eckrich \textbar{} OCHIN, Inc. \emph{Marion Ruth Sills, PI} \textbar{} Seattle Children's Hospital \emph{Dimitri A. Christakis, PI}, Daksha Ranade \textbar{} University of Missouri School of Medicine \emph{Abu Saleh Mohammad Mosa, PI}, Xing Song \textbar{} University of Texas Southwestern Medical Center \emph{Lindsay G. Cowell, PI} \textbar{} Weill Cornell Medicine \emph{Rainu Kaushal, mPI}, Thomas Campion

Authorship has been determined according to ICMJE recommendations.

We would like to thank the National Community Engagement Group (NCEG); all patients, caregivers, and community representatives; and all participants enrolled in the RECOVER Initiative. We would like to thank the patient representative, Elizabeth Noriega, for her contributions to this manuscript.

This content is solely the responsibility of the authors and does not necessarily represent the official views of the RECOVER Program, the NIH, or other funders.

\section*{Contributions}

All authors have read and approved the manuscript. Z.B.: Investigation, Writing, Formal analysis, Methodology; Z.X.: Investigation, Writing, Formal analysis; C.S.: Investigation, Writing; C.Z.: Investigation, Writing; H.T.B.: Investigation, Writing; C.S.: Investigation, Writing; J.R.: Investigation, Writing; A.T. M: Investigation, Writing; C.C.B.: Investigation, Writing; M.W.: Investigation, Writing; T.R.C.: Investigation, Writing; T.C.: Investigation, Writing; C.B.F.: Investigation, Writing; R.K.: Investigation, Writing, Funding acquisition, Resources; F.W.: Investigation, Writing, Supervision; Y.P.: Investigation, Writing, Supervision, Conceptualisation.

\section*{Competing interests}

No authors have any financial or non-financial competing interests.

\setlength{\bibsep}{2pt plus 0.3ex}
\bibliographystyle{unsrtnat}
\bibliography{sample}
%\printbibliography

\newpage
\appendix
\setcounter{subsection}{0}
\setcounter{table}{0}
\setcounter{figure}{0}
\renewcommand{\thesubsection}{S\arabic{subsection}}
\renewcommand\figurename{Supplementary Figure} 
\renewcommand\tablename{Supplementary Table}

% \section*{Supplementary materials}
% \label{sec:appendix}
% 
% {
% \centering
% \captionsetup{type=figure}
% \includegraphics{example-image-16x9}
% \captionof{figure}{caption}
% \label{fig2}}

\newpage

\end{document}

%% file: tables/table1.tex
\begin{tabularx}{\textwidth}{lrrrrX}
\toprule
\textbf{Dataset} & \textbf{Patients} & \textbf{Notes} & \textbf{Sentences} & \textbf{Mentions} & \textbf{Sites} \\
\midrule
Model Development Dataset & & & & & \\
\hspace{1em}WCM Training & 30 & 30 & 642 & 1,324 & WCM \\
\hspace{1em}2010 i2b2 assertion dataset\textsuperscript{28} & - & - & 3,055 & 4,243 & Partners Healthcare, Beth Israel Deaconess Medical Center, University of Pittsburgh Medical Center \\
\hspace{1em}Internal Validation & 30 & 30 & 350 & 953 & WCM \\
\hspace{1em}Multi-site External Validation & 100 & 100 & 1,113 & 1,886 & Medical College of Wisconsin, Cincinnati Children's Hospital Medical Center, The Children's Hospital of Philadelphia, University of Missouri, Nationwide Children's Hospital, Nemours Children's Health System, Oregon Community Health Information Network, Seattle Children's, UT Southwestern Medical Center, and Montefiore Medical Center \\
\midrule
Population-level Prevalence Study & 47,654 & 47,654 & - & - & Weill Cornell Medicine, Medical College of Wisconsin, Cincinnati Children's Hospital Medical Center, The Children's Hospital of Philadelphia, University of Missouri, Nationwide Children's Hospital, Nemours Children's Health System, Oregon Community Health Information Network, Seattle Children's, UT Southwestern Medical Center, and Montefiore Medical Center \\
\bottomrule
\end{tabularx}

%% file: tables/table2.tex
\begin{tabularx}{\textwidth}{lX}
\toprule
\textbf{Model} & \textbf{Hyperparameters} \\
\midrule
\textbf{BioBERT} & \begin{minipage}[t]{\linewidth}\raggedright
The following hyperparameters were used during training (provided on HuggingFace):

\begin{itemize}[nosep]
\item learning\_rate: 2e-05
\item train\_batch\_size: 8
\item eval\_batch\_size: 8
\item seed: 42
\item optimizer: Adam with betas=(0.9,0.999) and epsilon=1e-08
\item lr\_scheduler\_type: linear
\item num\_epochs: 10
\end{itemize}
\end{minipage} \\
\midrule
\textbf{ClinicalBERT} & \begin{minipage}[t]{\linewidth}\raggedright
The following hyperparameters were used during training (provided on HuggingFace):

\begin{itemize}[nosep]
\item batch\_size: 32
\item Maximum sequence length: 256
\item Learning rate: 5e-5
\end{itemize}
\end{minipage} \\
\midrule
\textbf{BiomedBERT} & \begin{minipage}[t]{\linewidth}\raggedright
The following hyperparameters were used during training (provided in the paper):

\begin{itemize}[nosep]
\item Optimizer: Adam
\item Learning Rate Schedule: Slanted triangular learning rate schedule with warm-up in 10\% of steps and cool-down in 90\% of steps.
\item Peak Learning Rate: $6\times 10^{-4}$
\item Training Steps: 62,500 steps
\item Batch Size: 8,192
\item Masking Rate for Whole-Word Masking (WWM): 15\%
\end{itemize}
\end{minipage} \\
\bottomrule
\end{tabularx}

%% file: tables/table3.tex
\begin{tabular}{rrrrrrrrrrrr}
\toprule
 & \textbf{cchmc} & \textbf{chop} & \textbf{missouri} & \textbf{monte} & \textbf{nationwide} & \textbf{nemours} & \textbf{seattle} & \textbf{ochin} & \textbf{utsw} & \textbf{mcw} & \textbf{wcm} \\
\midrule
\multicolumn{1}{l}{\textbf{Age} (mean ± SD)}  & 20.2 ± 4.4 & 18.9 ± 2.6 & 56.6 ± 18.7 & 58.0 ± 16.2  & 25.0 ± 11.1 & 18.6 ± 2.0 & 19.0 ± 3.4 & 50.5 ± 15.9 & 60.5 ± 15.9 & 57.6 ± 16.1  & 57.3 ± 18.1 \\
{[}95\% CI{]} & (20.1, 20.3) & (18.8, 19.0) & (56.1, 57.2) & (57.6, 58.5) & (24.7, 25.4) &  (18.5, 18.7) & (18.9, 19.2) & (50.1, 51.0) & (60.1, 60.9) & (57.1, 58.0) & (56.7, 57.8) \\
\midrule
\multicolumn{12}{l}{\textbf{Race}} \\
\makecell[rt]{American Indian \\or Alaska Native} & \makecell[rt]{8\\(0.2\%)} & \makecell[rt]{2\\(0.0\%)} & \makecell[rt]{15\\(0.3\%)} & \makecell[rt]{12\\(0.2\%)} & \makecell[rt]{3\\(0.1\%)} & \makecell[rt]{2\\(0.1\%)} & \makecell[rt]{13\\(0.7\%)} & \makecell[rt]{87\\(1.8\%)} & \makecell[rt]{16\\(0.3\%)} & \makecell[rt]{6\\(0.1\%)} & \makecell[rt]{9\\(0.2\%)} \\
Asian & \makecell[rt]{75\\(1.5\%)} & \makecell[rt]{193\\(3.9\%)} & \makecell[rt]{26\\(0.5\%)} & \makecell[rt]{135\\(2.7\%)} & \makecell[rt]{72\\(2.0\%)} & \makecell[rt]{46\\(2.2\%)} & \makecell[rt]{113\\(6.1\%)} & \makecell[rt]{436\\(9.2\%)} & \makecell[rt]{201\\(4.0\%)} & \makecell[rt]{56\\(1.1\%)} & \makecell[rt]{406\\(9.4\%)} \\
\makecell[rt]{Black \\or African American} & \makecell[rt]{900\\(18.0\%)} & \makecell[rt]{1308\\(26.2\%)} & \makecell[rt]{565\\(11.5\%)} & \makecell[rt]{1766\\(35.3\%)} & \makecell[rt]{962\\(26.3\%)} & \makecell[rt]{372\\(17.5\%)} & \makecell[rt]{121\\(6.6\%)} & \makecell[rt]{771\\(16.3\%)} & \makecell[rt]{868\\(17.4\%)} & \makecell[rt]{929\\(18.6\%)} & \makecell[rt]{634\\(14.7\%)} \\
\makecell[rt]{Native Hawaiian \\or Other Pacific Islander} & \makecell[rt]{6\\(0.1\%)} & \makecell[rt]{2\\(0.0\%)} & \makecell[rt]{2\\(0.0\%)} & \makecell[rt]{16\\(0.3\%)} & \makecell[rt]{9\\(0.2\%)} & \makecell[rt]{0\\(0.0\%)} & \makecell[rt]{17\\(0.9\%)} & \makecell[rt]{30\\(0.6\%)} & \makecell[rt]{6\\(0.1\%)} & \makecell[rt]{6\\(0.1\%)} & \makecell[rt]{2\\(0.0\%)} \\
White & \makecell[rt]{3746\\(74.9\%)} & \makecell[rt]{2904\\(58.1\%)} & \makecell[rt]{4207\\(85.6\%)} & \makecell[rt]{567\\(11.3\%)} & \makecell[rt]{2308\\(63.1\%)} & \makecell[rt]{1215\\(57.0\%)} & \makecell[rt]{1128\\(61.3\%)} & \makecell[rt]{2814\\(59.4\%)} & \makecell[rt]{3455\\(69.2\%)} & \makecell[rt]{3823\\(76.5\%)} & \makecell[rt]{2125\\(49.2\%)} \\
Multiple races & \makecell[rt]{126\\(2.5\%)} & \makecell[rt]{132\\(2.6\%)} & \makecell[rt]{0\\(0.0\%)} & \makecell[rt]{0\\(0.0\%)} & \makecell[rt]{184\\(5.0\%)} & \makecell[rt]{66\\(3.1\%)} & \makecell[rt]{85\\(4.6\%)} & \makecell[rt]{65\\(1.4\%)} & \makecell[rt]{0\\(0.0\%)} & \makecell[rt]{23\\(0.5\%)} & \makecell[rt]{0\\(0.0\%)} \\
Refuse to answer & \makecell[rt]{12\\(0.2\%)} & \makecell[rt]{0\\(0.0\%)} & \makecell[rt]{8\\(0.2\%)} & \makecell[rt]{279\\(5.6\%)} & \makecell[rt]{9\\(0.2\%)} & \makecell[rt]{20\\(0.9\%)} & \makecell[rt]{89\\(4.8\%)} & \makecell[rt]{128\\(2.7\%)} & \makecell[rt]{0\\(0.0\%)} & \makecell[rt]{11\\(0.2\%)} & \makecell[rt]{0\\(0.0\%)} \\
No information & \makecell[rt]{0\\(0.0\%)} & \makecell[rt]{3\\(0.1\%)} & \makecell[rt]{0\\(0.0\%)} & \makecell[rt]{389\\(7.8\%)} & \makecell[rt]{0\\(0.0\%)} & \makecell[rt]{0\\(0.0\%)} & \makecell[rt]{0\\(0.0\%)} & \makecell[rt]{5\\(0.1\%)} & \makecell[rt]{30\\(0.6\%)} & \makecell[rt]{1\\(0.0\%)} & \makecell[rt]{1080\\(25.0\%)} \\
Other & \makecell[rt]{127\\(2.5\%)} & \makecell[rt]{0\\(0.0\%)} & \makecell[rt]{84\\(1.7\%)} & \makecell[rt]{1836\\(36.7\%)} & \makecell[rt]{0\\(0.0\%)} & \makecell[rt]{394\\(18.5\%)} & \makecell[rt]{275\\(14.9\%)} & \makecell[rt]{39\\(0.8\%)} & \makecell[rt]{99\\(2.0\%)} & \makecell[rt]{140\\(2.8\%)} & \makecell[rt]{0\\(0.0\%)} \\
Unknown & \makecell[rt]{0\\(0.0\%)} & \makecell[rt]{456\\(9.1\%)} & \makecell[rt]{8\\(0.2\%)} & \makecell[rt]{0\\(0.0\%)} & \makecell[rt]{111\\(3.0\%)} & \makecell[rt]{15\\(0.7\%)} & \makecell[rt]{0\\(0.0\%)} & \makecell[rt]{366\\(7.7\%)} & \makecell[rt]{320\\(6.4\%)} & \makecell[rt]{5\\(0.1\%)} & \makecell[rt]{64\\(1.5\%)} \\
\midrule
\multicolumn{12}{l}{\textbf{Ethnicity}} \\
Hispanic & \makecell[rt]{173\\(3.5\%)} & \makecell[rt]{445\\(8.9\%)} & \makecell[rt]{91\\(1.9\%)} & \makecell[rt]{2115\\(42.3\%)} & \makecell[rt]{205\\(5.6\%)} & \makecell[rt]{467\\(21.9\%)} & \makecell[rt]{311\\(16.9\%)} & \makecell[rt]{1723\\(36.3\%)} & \makecell[rt]{570\\(11.4\%)} & \makecell[rt]{168\\(3.4\%)} & \makecell[rt]{669\\(15.5\%)} \\
Non-Hispanic & \makecell[rt]{4799\\(96.0\%)} & \makecell[rt]{4517\\(90.3\%)} & \makecell[rt]{4801\\(97.7\%)} & \makecell[rt]{2493\\(49.9\%)} & \makecell[rt]{3426\\(93.7\%)} & \makecell[rt]{1623\\(76.2\%)} & \makecell[rt]{1440\\(78.2\%)} & \makecell[rt]{2802\\(59.1\%)} & \makecell[rt]{4248\\(85.0\%)} & \makecell[rt]{4815\\(96.3\%)} & \makecell[rt]{3651\\(84.5\%)} \\
\midrule
\multicolumn{12}{l}{\textbf{Sex}} \\
Male & \makecell[rt]{2118\\(42.4\%)} & \makecell[rt]{2088\\(41.8\%)} & \makecell[rt]{2249\\(45.8\%)} & \makecell[rt]{1556\\(31.1\%)} & \makecell[rt]{1237\\(33.8\%)} & \makecell[rt]{868\\(40.8\%)} & \makecell[rt]{778\\(42.3\%)} & \makecell[rt]{1577\\(33.3\%)} & \makecell[rt]{2097\\(42.0\%)} & \makecell[rt]{1967\\(39.3\%)} & \makecell[rt]{1711\\(39.6\%)} \\
Female & \makecell[rt]{2882\\(57.6\%)} & \makecell[rt]{2912\\(58.2\%)} & \makecell[rt]{2666\\(54.2\%)} & \makecell[rt]{3444\\(68.9\%)} & \makecell[rt]{2421\\(66.2\%)} & \makecell[rt]{1262\\(59.2\%)} & \makecell[rt]{1060\\(57.6\%)} & \makecell[rt]{3155\\(66.5\%)} & \makecell[rt]{2898\\(58.0\%)} & \makecell[rt]{3033\\(60.7\%)} & \makecell[rt]{2609\\(60.4\%)} \\
\bottomrule
\end{tabular}